\documentclass[conference]{IEEEtran}
\IEEEoverridecommandlockouts
\usepackage{cite}
\usepackage{amsmath,amssymb,amsfonts}
\usepackage{algorithmic}
\usepackage{graphicx}
\usepackage{textcomp}
\usepackage[hyphens]{url}
\usepackage[table,xcdraw]{xcolor}
\usepackage{multirow}
\usepackage{subfigure}
\usepackage{threeparttable}
\usepackage{hyperref}
\def\BibTeX{{\rm B\kern-.05em{\sc i\kern-.025em b}\kern-.08em
    T\kern-.1667em\lower.7ex\hbox{E}\kern-.125emX}}
\begin{document}

\title{Merging and Evolution: Improving Convolutional Neural Networks for Mobile Applications}

\def\todo#1{\textcolor{red}{#1}}
\newcolumntype{P}[1]{>{\centering\arraybackslash}p{#1}}
\newcolumntype{L}[1]{>{\arraybackslash}p{#1}}

\author{
\IEEEauthorblockN{Zheng Qin, Zhaoning Zhang$^{*}$\thanks{$^*$Corresponding author.}, Shiqing Zhang, Hao Yu, Yuxing Peng}
\IEEEauthorblockA{\textit{Science and Technology on Parallel and Distributed Laboratory}\\
\textit{National University of Defense Technology}\\
Changsha, China\\
qinzheng12@nudt.edu.cn; zzningxp@gmail.com; \{zhangshiqing12, yuhao12, pengyuxing\}@nudt.edu.cn}
}

\hyphenation{networks methods para-meters appears op-tical net-works semi-conduc-tor store stor-age snap-shot space reads among Ursa exits scal-able uses sys-tems IOPS dif-ferent GFS HDFS FS gen-erally SCSI NFS HBase SSD slightly stored two anony-mized speak-ing lineariza-bility server elim-inate states pro-cesses signifi-cantly sin-gle OCFS blocks par-allelism parallel-ism returns pre-senting present-ing stand-ard stores IOPS servers appends scala-bility pri-mary mode respec-tively re-spectively sat-isfy UPSs CPUs avail-ability availa-bility availabil-ity clouds usu-ally against re-quests inde-pendently namely storing Meituan evolv-ing since ob-ject tests BS mounted under-loaded consis-tency status jour-nals place-ment statis-tics exclu-sively meta-data mainly collabora-tively collabo-ratively FDS focus nodes remote conven-tional strata single enough daily require small sequen-tial Theore-tically coroutine acce-leration accele-ration effi-cient matrix vectors imple-mentations implemen-tations implementa-tions larger Mobile-Nets GPUs com-puted cuDNN hyper-parameters hyperpara-meters Con-sequently Conse-quently app-lications appli-cations applica-tions adopted GPU accele-rating acce-lerating Tensor-Flow results frame-works convo-lutions con-volutions convolu-tions convo-lution con-volution convolu-tion threads signi-ficant signifi-cant sig-nificant ii iii GEMM aca-demic acade-mic its weights algo-rithms uti-lize also rela-tively redun-dancy imple-mentation implemen-tation implementa-tion Mobile-Net multi-plier fewer res-pectively respec-tively kernel table uti-lizes every in-creases diag-onalwise Multiplications be-nefit be-nefits mo-dification suf-fer in-duced shu-ffle evo-lution mo-dules alle-viating discrimi-native com-putational ope-rations mo-dule fa-mily fami-ly fur-ther ope-ration chan-nels}

\hyphenpenalty=1000

\maketitle

\begin{abstract}
Compact neural networks are inclined to exploit ``sparsely-connected'' convolutions such as depthwise convolution and group convolution for employment in mobile applications.
Compared with standard ``fully-connected'' convolutions, these convolutions are more computationally economical.
However, ``sparsely-connected'' convolutions block the inter-group information exchange, which induces severe performance degradation.
To address this issue, we present two novel operations named \emph{merging} and \emph{evolution} to leverage the inter-group information.
Our key idea is encoding the inter-group information with a narrow feature map, then combining the generated features with the original network for better representation.
Taking advantage of the proposed operations, we then introduce the \emph{Merging-and-Evolution (ME) module}, an architectural unit specifically designed for compact networks.
Finally, we propose a family of compact neural networks called \emph{MENet} based on ME modules.
Extensive experiments on ILSVRC 2012 dataset and PASCAL VOC 2007 dataset demonstrate that MENet consistently outperforms other state-of-the-art compact networks under different computational budgets.
For instance, under the computational budget of 140 MFLOPs, MENet surpasses ShuffleNet by 1\% and MobileNet by 1.95\% on ILSVRC 2012 top-1 accuracy, while by 2.3\% and 4.1\% on PASCAL VOC 2007 mAP, respectively.
\end{abstract}

\begin{IEEEkeywords}
Convolutional neural networks;
deep learning;
model acceleration
\end{IEEEkeywords}

\section{Introduction}
\label{section:introduction}

Convolutional neural networks (CNNs) have achieved significant progress in computer vision tasks such as image classification \cite{krizhevsky2012imagenet, simonyan2014very, he2016deep, szegedy2015going, szegedy2017inception}, object detection \cite{ren2015faster, liu2016ssd, redmon2016yolo9000, he2017mask} and semantic segmentation \cite{long2015fully}.
However, state-of-the-art CNNs require computation at billions of FLOPs, which prevents them from being utilized in mobile or embedded applications.
For instance, ResNet-101 \cite{he2016deep}, which is broadly used in detection tasks \cite{ren2015faster, he2017mask}, has a complexity of 7.8 GFLOPs and fails to achieve real-time detection even with a powerful GPU.

\begin{figure}[!t]
\centering
\includegraphics[width=0.45\textwidth]{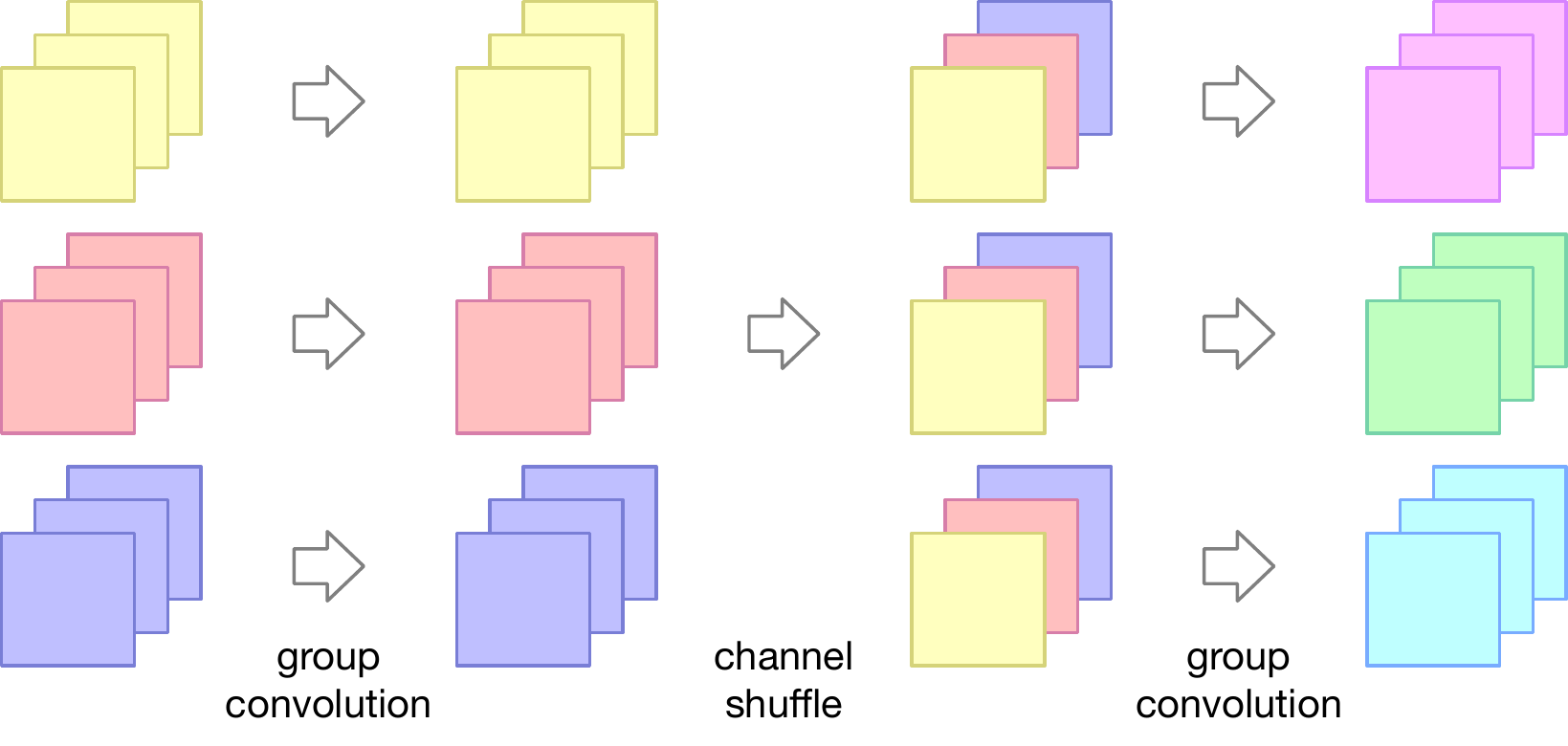}
\caption{Channel shuffle operation with 9 channels and 3 channel groups. Each group in the second convolution receives only 1 channel from each group in the first convolution. This leads to severe inter-group information loss.}
\label{figure:channel-shuffle}
\end{figure}

In view of the huge computational cost of modern CNNs, compact neural networks \cite{iandola2016squeezenet, howard2017mobilenets, zhang2017shufflenet} have been proposed to deploy both accurate and efficient networks on mobile or embedded devices.
Compact networks can achieve relatively high accuracy under a tight computational budget.
For better computational efficiency, these networks are inclined to utilize ``sparsely-connected'' convolutions such as depthwise convolution and group convolution rather than standard ``fully-connected'' convolutions.
For instance, ShuffleNet \cite{zhang2017shufflenet} utilizes a lightweight version of the bottleneck unit \cite{he2016deep} termed ShuffleNet unit.
In a ShuffleNet unit, the original $3 \times 3$ convolution is replaced with a $3 \times 3$ depthwise convolution, while the $1 \times 1$ convolutions are substituted with pointwise group convolutions.
This modification significantly reduces the computational cost, but blocks the information flow between channel groups and leads to severe performance degradation.
For this reason, ShuffleNet introduces the channel shuffle operation to enable inter-group information exchange.
As shown in Fig.~\ref{figure:channel-shuffle}, a channel shuffle operation permutes the channels so each group in the second convolutional layer contains channels from every group in the first convolutional layer.
Benefiting from the channel shuffle operation, ShuffleNet achieves 65.9\% top-1 accuracy on ILSVRC 2012 dataset \cite{russakovsky2015imagenet} with 140 MFLOPs, and 70.9\% top-1 accuracy with 524 MFLOPs, which is state-of-the-art.

However, the channel shuffle operation fails to eliminate the performance degradation and ShuffleNet still suffers from the loss of inter-group information.
Fig.~\ref{figure:channel-shuffle} illustrates a channel shuffle operation with 9 channels and 3 channel groups.
Each group in the second convolutional layer receives only 1 channel from every group in the first convolutional layer, whereas there are 2 other channels in each group being ignored.
As a result, a large portion of the inter-group information cannot be leveraged.
This problem is aggravated given more channel groups.
Although there are more channels in total given more groups, the number of channels in each group is smaller, which increases the loss of inter-group information.
Consequently, when the computational budget is relatively larger, ShuffleNet architectures with more channel groups perform worse than the narrower ones which have less groups.
This indicates that it is difficult for ShuffleNet to gain performance increase by increasing the number of channels directly.



To address this issue, we propose two novel operations named \emph{merging} and \emph{evolution} to directly fuse features across all channels in a group convolution and alleviate the loss of inter-group information.
For a feature map generated from a group convolution, a merging operation aggregates the features at the same spatial position across all channels and encodes the inter-group information into a narrow feature map.
An evolution operation is performed afterwards to extract spatial information from the feature map.
Then, based on the proposed operations, we introduce the \emph{Merging-and-Evolution (ME) module}, a powerful and efficient architectural unit specifically for compact networks.
For computational efficiency, ME modules exploit depthwise convolutions and group convolutions to reduce the computational cost.
For better representation, ME modules utilize merging and evolution operations to leverage the inter-group information.
Finally, we present a new family of compact neural networks called \emph{MENet} which is built with ME modules.
Compared with ShuffleNet \cite{zhang2017shufflenet}, MENet alleviates the loss of inter-group information and gains substantial improvements as the group number increases.

We conduct extensive experiments to evaluate the effectiveness of MENet.
Firstly, we compare MENet with other state-of-the-art network structures on the ILSVRC 2012 classification dataset \cite{russakovsky2015imagenet}.
Then, we examine the generalization ability of MENet on the PASCAL VOC 2007 detection dataset \cite{everingham2010pascal}.
Experiments show that MENet consistently outperforms other state-of-the-art compact networks under different computational budgets.
For instance, under a complexity of 140 MFLOPs, MENet achieves improvements of 1\% over ShuffleNet and 1.95\% over MobileNet on ILSVRC 2012 top-1 accuracy, while 2.3\% and 4.1\% on PASCAL VOC 2007 mAP, respectively.
Our models have been made publicly available at \texttt{\url{https://github.com/clavichord93/MENet}}.

\section{Related Work}

As deep neural networks suffer from heavy computational cost and large model size, the inference-time compression and acceleration of neural networks has become an attractive topic in deep learning community.
Commonly, the related work can be categorized into four groups.

\emph{Tensor decomposition} factorizes a convolution into a sequence of smaller convolutions with fewer parameters and less computational cost.
Jaderberg \textit{et al.} \cite{jaderberg2014speeding} proposed to decompose a $k \times k$ convolution into a $k \times 1$ convolution and a $1 \times k$ convolution, reporting $4.5 \times$ speedup with 1\% accuracy loss.
Denton \textit{et al.} \cite{denton2014exploiting} proposed a method exploiting a low-rank decomposition to estimate the original convolution.
Recently, Zhang \textit{et al.} \cite{zhang2016accelerating} proposed a method based on generalized singular value decomposition without the need of stochastic gradient descent, which achieved $4 \times$ speedup on VGG-16 \cite{simonyan2014very} with a graceful accuracy degradation.

\emph{Parameter quantization} is proposed to utilize low-bit parameters in neural networks.
Vanhoucke \textit{et al.} \cite{vanhoucke2011improving} proposed to use 8-bit fixed-point parameters and achieved $3 \times$ speedup.
Gong \textit{et al.} \cite{gong2014compressing} applied k-means clustering on network parameters and provided $20 \times$ compression with only 1\% accuracy drop.
Binarization methods \cite{courbariaux2016binarized, zhou2016dorefa, rastegari2016xnor} attempted to train networks directly with 1-bit weights.
Quantization methods provide significant memory savings and enormous theoretical speedup.
However, current hardware is mainly optimized for half-/single-/double-precision computation, so it is difficult for quantization methods to achieve the theoretical speedup.

\emph{Network pruning} attempts to recognize the structure redundancy in network architectures and cut off the redundant parameters.
Han \textit{et al.} \cite{han2015learning} proposed a method to remove all connections with small weights, reporting $10 \times$ reduction in model size.
Network slimming \cite{liu2017learning} applied sparsity-induced penalty on the scaling factors in batch normalization layers and removes the channels with small scaling factors.
He \textit{et al.} \cite{he2017channel} proposed a LASSO regression based method to prune redundant channels, achieving $5 \times$ speedup with comparable accuracy.
Yu \textit{et al.} \cite{yu2017accelerating} proposed a group-wise 2D-filter pruning approach and provided $4 \times$ speedup on VGG-16.
However, iterative pruning strategy is commonly utilized in network pruning, which slows down the training procedure.

\emph{Compact networks} are designed for mobile or embedded applications specifically.
SqueezeNet \cite{iandola2016squeezenet} proposed fire modules, where a $1 \times 1$ convolutional layer is first applied to ``squeeze'' the width of the network, followed by a layer mixing $3 \times 3$ and $1 \times 1$ convolutional kernels to reduce parameters.
MobileNet \cite{howard2017mobilenets} exploited depthwise separable convolutions as its building unit, which decompose a standard convolution into a combination of a depthwise convolution and a pointwise convolution.
ShuffleNet \cite{zhang2017shufflenet} utilized depthwise convolutions and pointwise group convolutions into the bottleneck unit \cite{he2016deep}, and proposed the channel shuffle operation to enable inter-group information exchange.
Compact networks can be trained from the scratch, so the training procedure is very fast.
Moreover, compact networks are orthogonal to the aforementioned methods and can be further compressed.


\section{Merging-and-Evolution Networks}

In this section, we first analyze the loss of inter-group information in ShuffleNet and introduce merging and evolution operations for alleviating the performance degradation.
Next, we describe the structure of the ME module.
At last, the details about MENet architecture are introduced.

\subsection{Merging and Evolution Operations}
\label{section:merging-evolution-operation}

As figured out in Section~\ref{section:introduction}, ShuffleNet suffers from the severe inter-group information loss.
The loss of inter-group information can be measured with the number of inter-group connections.
Specifically, for two consecutive convolutional layers with $C$ output channels and $G$ channel groups, each group contains $\frac{C}{G}$ channels, and there are totally
\begin{equation}
N_{total}=\frac{1}{2}\cdot C \cdot(C-\frac{C}{G}) = \frac{C^2(G-1)}{2G}
\end{equation}
inter-group connections if the channels were ``fully-connected''.
After a channel shuffle operation, each group in the later convolutional layer receives $\frac{C}{G^2}$ channels from every group in the former layer, so there are
\begin{equation}
N_{actual}=\frac{1}{2}\cdot C \cdot(\frac{C}{G}-\frac{C}{G^2}) = \frac{C^2(G-1)}{2G^2}
\end{equation}
actual inter-group connections.
This means a ratio of 
\begin{equation}
1 - \frac{N_{actual}}{N_{total}} = 1 - \frac{C^2(G-1) / (2G^2)}{C^2(G-1) / (2G)} = \frac{G-1}{G}
\end{equation}
of the inter-group connections are lost, which induces severe loss of inter-group information.
This significantly weakens the representation capability and leads to serious performance degradation.
The problem is aggravated when there are more channel groups.
The ratio of the inter-group connections lost is 66.7\% when there are three groups, but the number increases to 87.5\% given eight groups.
This explains why ShuffleNet with three groups outperforms the one with eight groups.

\begin{figure}[!t]
\centering
\includegraphics[width=0.48\textwidth]{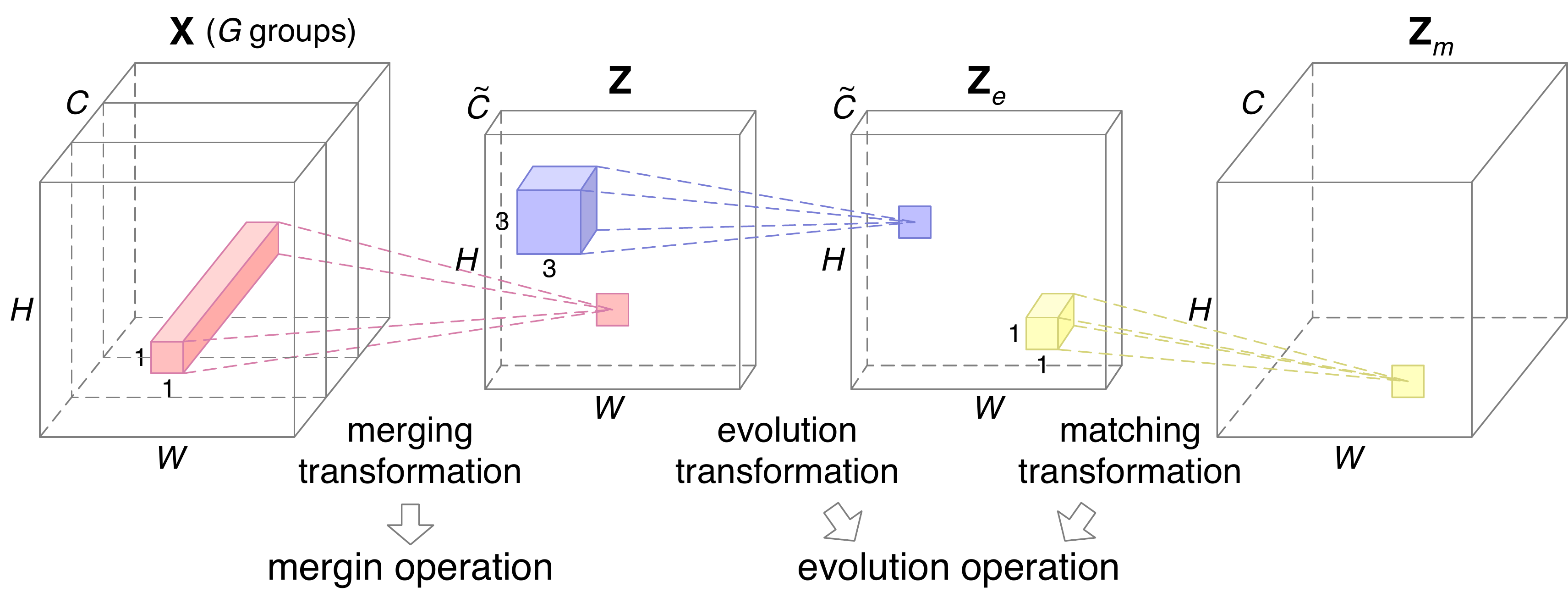}
\caption{Merging and evolution operations. A merging operation applies a merging transformation and encodes the inter-group information into a narrow feature map. An evolution operation consists of an evolution transformation and a matching transformation and leverages spatial information.}
\label{figure:ME-operation}
\end{figure}

To address this issue, we design two operations termed \emph{merging} and \emph{evolution} to leverage inter-group information.
As shown in Fig.~\ref{figure:ME-operation}, the proposed operations encode the inter-group information with a narrow feature map, and combine it with the original network for more discriminative features.

\subsubsection{Merging Operation}


The merging operation is designed to fuse features across all channels and encode the inter-group information into a narrow feature map.
Given the feature map $\textbf{X} \in \mathbb{R}^{C \times H \times W}$ generated from a group convolution, a merging transformation $f: \mathbb{R}^{C \times H \times W} \rightarrow \mathbb{R}^{\tilde{C} \times H \times W}$ is applied to aggregate features over all channels, where $C$ is the number of channels in the original feature map, $H$ and $W$ are the spatial dimensions, and $\tilde{C}$ is the number of channels in the produced feature map.
A small $\tilde{C}$ is chosen to make the merging operations computationally economical.
As $C$ is relatively large, it is difficult to integrate the spatial information without harming the computational efficiency.
So we aggregate only the features on the same spatial position along all channels in a merging operation.
A single pointwise convolution is exploited as the merging transformation, followed by a batch normalization \cite{ioffe2015batch} and a ReLU activation.
Formally, the output feature map of a merging operation is calculated as
\begin{equation}
\textbf{Z} = \delta(\text{BN}(f(\textbf{X}))) = \delta(\text{BN}(\textbf{W} \ast \textbf{X}))
\end{equation}
where $\delta$ indicates the ReLU function, $\ast$ represents the convolution operator, and $\textbf{W} \in \mathbb{R}^{\tilde{C} \times C \times 1 \times 1}$ is the convolutional kernel.
By this means, each channel in $\textbf{Z}$ contains information from every channel in the previous group convolutional layer.

\subsubsection{Evolution Operation}

After a merging operation, an evolution operation is performed to obtain more discriminative features.
An evolution operation is defined in two steps.
In the first step, an evolution transformation $g_e: \mathbb{R}^{\tilde{C} \times H \times W} \rightarrow \mathbb{R}^{\tilde{C} \times H \times W}$ is applied to the feature map from the previous merging operation.
The number of channels is kept unchanged.
In this step, we intend to leverage more spatial information so a $3 \times 3$ standard convolution is selected as the evolution transformation, followed by a batch normalization and a ReLU activation.
In the second step, a matching transformation $g_m: \mathbb{R}^{\tilde{C} \times H \times W} \rightarrow \mathbb{R}^{C \times H \times W}$ is performed to match the size of the output feature map with the original network.
As in the merging operation, a single pointwise convolution is chosen as $g_m$ to maintain the computational efficiency.
Another batch normalization and a sigmoid activation are added afterwards.
The whole process is formally written as
\begin{align}
\textbf{Z}_e & = \delta(\text{BN}(g_e(\textbf{Z}))) = \delta(\text{BN}(\textbf{W}_e \ast \textbf{Z})) \\
\textbf{Z}_m & = \sigma(\text{BN}(g_m(\textbf{Z}_e))) = \sigma(\text{BN}(\textbf{W}_m \ast \textbf{Z}_e)) \notag \\
& = \sigma(\text{BN}(\textbf{W}_m \ast \delta(\text{BN}(\textbf{W}_e \ast \textbf{Z}))))
\end{align}
where $\textbf{W}_e \in \mathbb{R}^{\tilde{C} \times \tilde{C} \times 3 \times 3}$ and $\textbf{W}_m \in \mathbb{R}^{C \times \tilde{C} \times 1 \times 1}$ are the convolutional kernels, and $\sigma$ indicates the sigmoid function.

At last, the features generated from evolution operations are regarded as \emph{neuron-wise scaling factors} and combined with the original network using an element-wise product to improve the representation capability of the features in the network:
\begin{align}
\tilde{\textbf{X}} & = h(\textbf{X}) \circ \textbf{Z}_m \notag \\
& = h(\textbf{X}) \circ \sigma(\text{BN}(\textbf{W}_m \ast \delta(\text{BN}(\textbf{W}_e \ast \textbf{Z}))))
\end{align}
where $h: \mathbb{R}^{C \times H \times W} \rightarrow \mathbb{R}^{C \times H \times W}$ is the transformation in the original network, and $\circ$ represents element-wise product.
As $\textbf{Z}_m$ encodes information from every channel in the previous convolution, each channel in $\tilde{\textbf{X}}$ also contains information from all channels.
This alleviates the loss of inter-group information.

\begin{figure*}[!t]
\centering
\subfigure[]{
\includegraphics[width=0.32\textwidth]{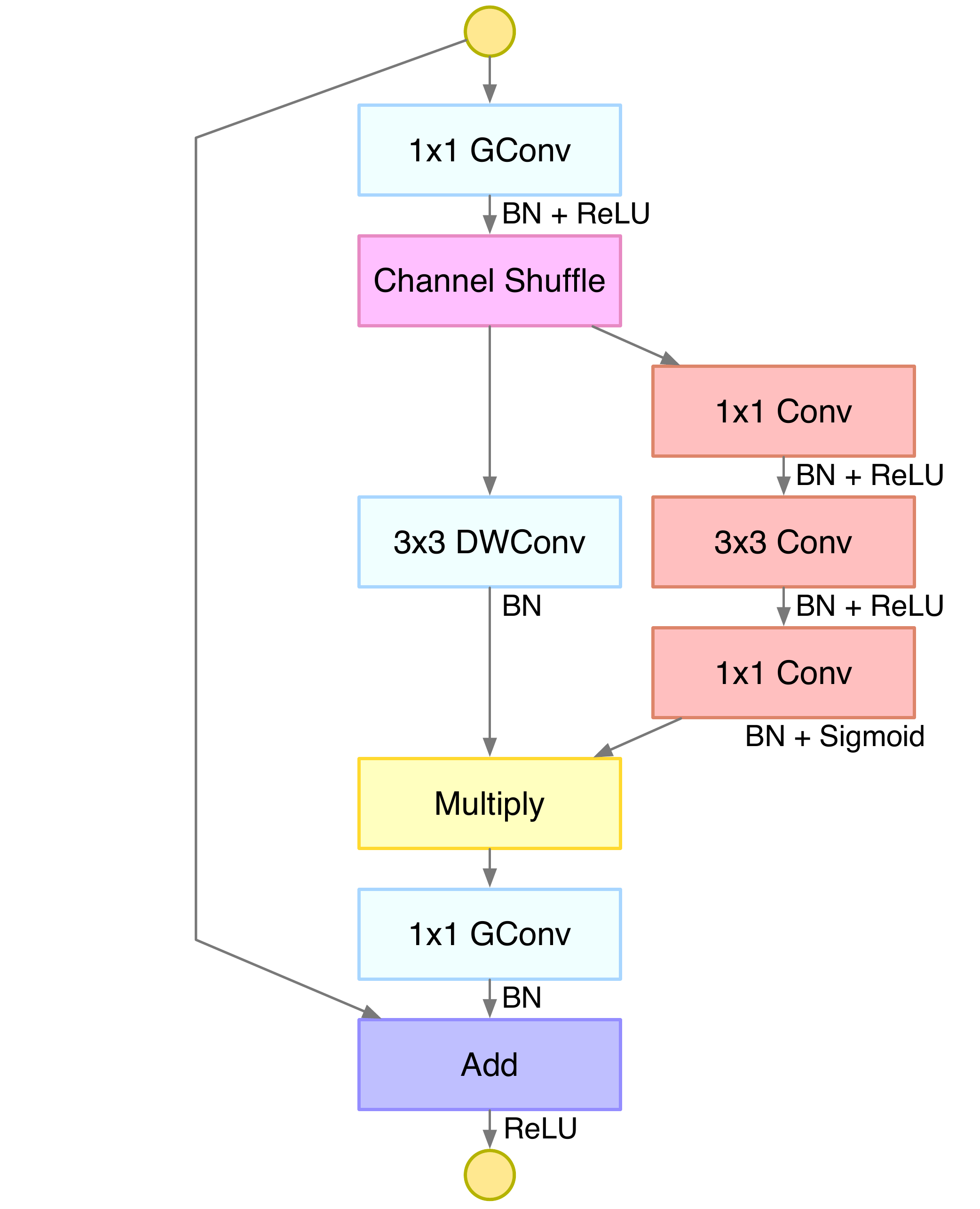}
\label{figure:ME-module-s1}
}
\subfigure[]{
\includegraphics[width=0.32\textwidth]{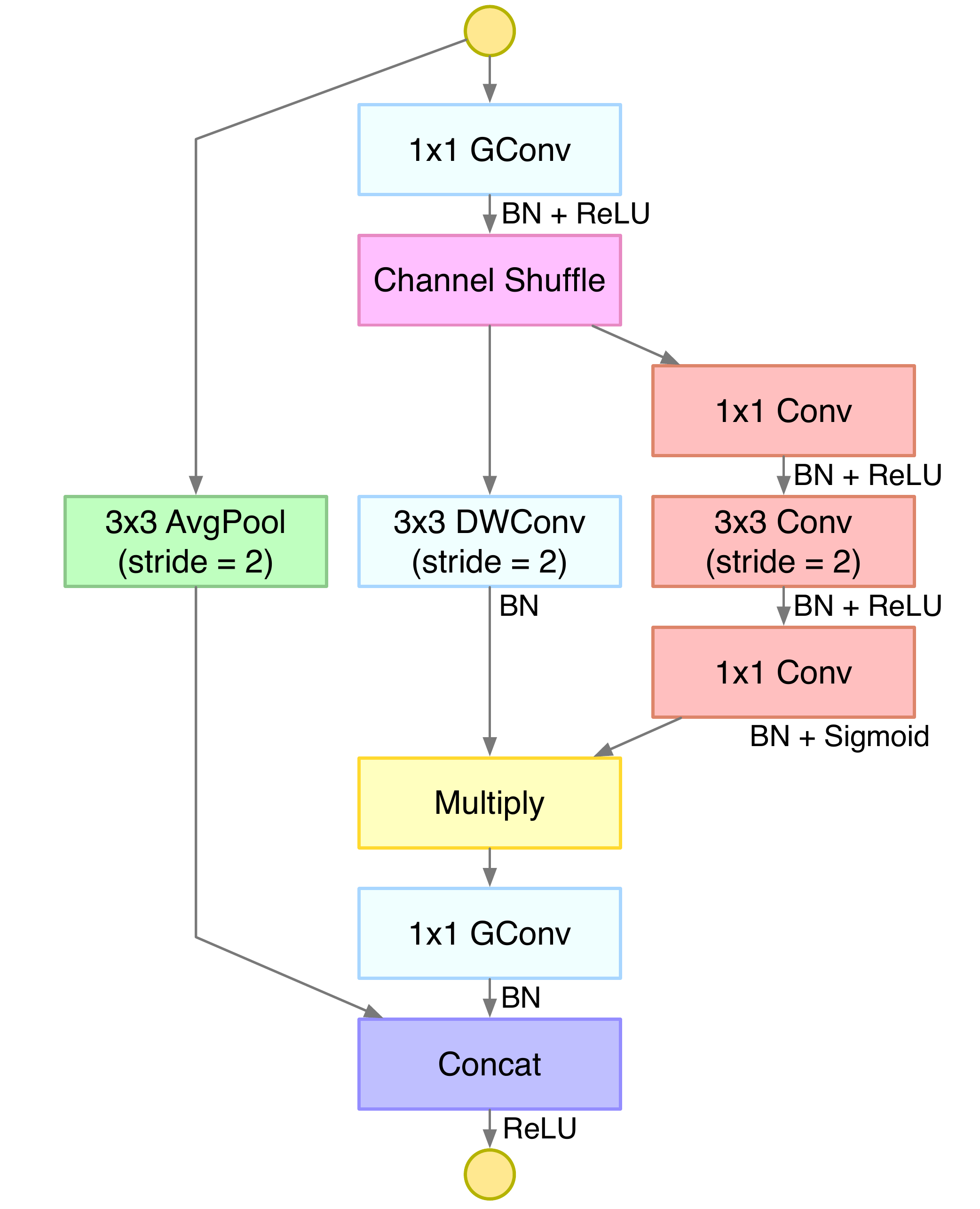}
\label{figure:ME-module-s2}
}
\caption{The structure of ME module. \textbf{(a)}: Standard ME module. \textbf{(b)}: Downsampling ME module. \textbf{GConv}: Group convolution. \textbf{DWConv}: Depthwise convolution.}
\label{figure:ME-module}
\end{figure*}

\subsection{Merging-and-Evolution Module}

Taking advantage of the proposed merging and evolution operations, we present the \emph{Merging-and-Evolution (ME) module}, an architectural unit specifically designed for compact neural networks.

The ME module is a variant of the conventional residual block \cite{he2016deep}.
An ME module consists of three branches: an \emph{identity branch}, a \emph{residual branch} and a \emph{fusion branch}, as illustrated from left to right in Fig.\ref{figure:ME-module-s1}.
For computational efficiency, the residual branch adopts a bottleneck design \cite{he2016deep} and exploits ``sparsely-connected'' convolutions.
It consists of three layers, a pointwise group convolution to squeeze the channel dimension, a $3 \times 3$ depthwise convolution to leverage spatial information, and another pointwise group convolution to recover the channel dimension.
A channel shuffle operation \cite{zhang2017shufflenet} is applied after the first pointwise group convolution for inter-group information exchange.
The utilization of merging and evolution operations introduces the fusion branch.
A merging operation is performed after the channel shuffle, with an evolution operation following.
Then the fusion branch is combined with the residual branch before the second pointwise group convolutional layer.
This design helps alleviate the loss of inter-group information in the second group convolutional layer.
The merging and evolution operations are applied to the bottleneck channels to reduce the overall computational cost.
Additionally, as described in Section~\ref{section:merging-evolution-operation}, the number of channels in the fusion branch is kept small to maintain computational efficiency.

For the downsampling version of ME modules, two more modifications are performed.
(i) The strides of the depthwise convolution in the residual branch and the $3 \times 3$ convolution in the fusion branch are altered to 2.
(ii) Inspired by \cite{zhang2017shufflenet}, a $3 \times 3$ average pooling with a stride of 2 is applied in the identity branch, and the element-wise addition is substituted with a concatenation to combine the identity branch and the residual branch.
After a downsampling ME module, the spatial dimensions of the feature map are halved, while the channel dimension is doubled.
Fig.~\ref{figure:ME-module-s2} describes the structure of the downsampling ME module.

\begin{table*}[!t]
\centering
\begin{threeparttable}[!t]
\caption{MENet Architecture for ImageNet under the Computational Budget of 140 MFLOPs}
\label{table:menet-architecture}
\begin{tabular}{|c|c|c|c|c|}
\hline
\textbf{Stage} & \textbf{Output Size}& \textbf{228-MENet-12$\times$1 ($g=3$)} & \textbf{256-MENet-12$\times$1 ($g=4$)} & \textbf{352-MENet-12$\times$1 ($g=8$)} \\ \hline
 & $224 \times 224$ & \multicolumn{3}{c|}{Image} \\ \hline
\multirow{2}{*}{Stage 1} & $112 \times 112$ & \multicolumn{3}{c|}{$3 \times 3$ conv, 24, /2} \\
 & $56 \times 56$ & \multicolumn{3}{c|}{$3 \times 3$ max pool, /2} \\ \hline
\multirow{2}{*}{Stage 2} & \multirow{2}{*}{$28 \times 28$} & ME module, 228, /2 & ME module, 256, /2 & ME module, 352, /2 \\
 & & ME module, 228, $\times$3 & ME module, 256, $\times$3 & ME module, 352, $\times$3 \\ \hline
\multirow{2}{*}{Stage 3} & \multirow{2}{*}{$14 \times 14$} & ME module, 456, /2 & ME module, 512, /2 & ME module, 704, /2 \\
 & & ME module, 456, $\times$7 & ME module, 512, $\times$7 & ME module, 704, $\times$7 \\ \hline
\multirow{2}{*}{Stage 4} & \multirow{2}{*}{$7 \times 7$} & ME module, 912, /2 & ME module, 1024, /2 & ME module, 1408, /2 \\
 & & ME module, 912, $\times$3 & ME module, 1024, $\times$3 & ME module, 1408, $\times$3 \\ \hline
Classifier & $1 \times 1$ & \multicolumn{3}{c|}{global average pool, 1000-d fc, softmax} \\ \hline
FLOPs &  & $144 \times 10^6$ & $140 \times 10^6$ & $144 \times 10^6$ \\ \hline
\end{tabular}
The number after the layer/module type is the number of output channels.
``$\times$3'' and ``$\times$7'' indicate the ME module repeats 3 or 7 times respectively.
``/2'' represents the stride of the layer is 2.
The ME modules with ``/2'' perform downsampling.
\end{threeparttable}
\end{table*}

\subsection{MENet Architecture}
\label{section:menet-architecture}

Based on ME modules, we propose \emph{MENet}, a new family of compact neural networks.
The overall architecture of MENet for ImageNet classification is demonstrated in Table~\ref{table:menet-architecture}.

MENet begins with a $3 \times 3$ convolutional layer and a max pooling layer, both with strides of 2.
A batch normalization and a ReLU activation are applied after the convolutional layer.
These two layers perform $4 \times$ downsampling to reduce the overall computational cost.
Then there follow a sequence of ME modules, which are grouped into three stages (Stage 2 to 4).
In each stage, the first building block is a downsampling ME module, while the rest building blocks are standard ME modules.
The number of output channels is kept the same within a stage and is doubled in the next stage.
Furthermore, the number of bottleneck channels in the residual branch is set to $1/4$ of the output channels in the same ME module, and we do not apply group convolution on the first pointwise layer in Stage 2.
We build MENet with three \emph{group numbers} $g$: $g = 3$, $g = 4$ and $g = 8$.
Increasing the group number aggravates the connection sparsity in the residual branch, but contributes to wider feature maps.
The influence of the group number on the performance of MENet is discussed in the next section.

We furthermore introduce three hyper-parameters for customizing MENet to fit different computational budgets.
The first two hyper-parameters are the \emph{fusion width} $k$ and the \emph{expansion factor} $\alpha$, which control the complexity of the fusion branch.
The fusion width is defined as the number of channels in the fusion branch of Stage 2, and the expansion factor represents the ratio of the channels in the fusion branch between two consecutive stages.
The number of channels in the fusion branch of Stage $i$ ($i\geq2$) is calculated as $\alpha^{i-2}k$.
We figure that intuitively it is beneficial for generating more discriminative features to have wider fusion branches, but it also leads to more computational cost.
The effects of the fusion width and the expansion factor on the performance of MENet is discussed in the next section.
The third hyper-parameter is the \emph{residual width} $w$, which is defined as the number of output channels in the residual branch of Stage 2.
The residual width controls the computational cost in the residual branch.


Finally, we define a notation ``$w$-MENet-$k \times \alpha$'' to represent a network with a residual width $w$, a fusion width $k$ and an expansion factor $\alpha$.
For example, the network in Table~\ref{table:menet-architecture} with $g = 3$ can be denoted as ``228-MENet-$12 \times 1$''.

\section{Experiments}

We conduct extensive experiments to examine the effectiveness of MENet with two benchmarks.
We first evaluate MENet on the ILSVRC 2012 classification dataset \cite{russakovsky2015imagenet} and compare MENet with other state-of-the-art networks.
The influence of different model choices is then investigated.
At last, we conduct experiments on the PASCAL VOC 2007 detection dataset \cite{everingham2010pascal} to examine the generalization ability of MENet.

\begin{table}[!t]
\centering
\setlength{\tabcolsep}{4.5pt}
\begin{threeparttable}[!t]
\caption{ILSVRC 2012 Accuracy (\%) Comparison with State-of-the-art Network Structures}
\label{table:result-state-of-the-art}
\begin{tabular}{|l|c|c|c|}
\hline
\textbf{Models} & \textbf{MFLOPs} & \textbf{Top-1 Acc.} & \textbf{Top-5 Acc.} \\ \hline
VGG-16 \cite{simonyan2014very} & 15300 & 71.5 & 89.8 \\
GoogLeNet \cite{szegedy2015going} & 1550 & 69.8 & 89.6 \\
456-MENet-24$\times$1 ($g=3$, \emph{ours}) & 551 & \textbf{71.6} & \textbf{90.2} \\ \hline
ResNet-15 \cite{he2016deep} & 140 & 61.3 & - \\
Xception-15 \cite{chollet2016xception} & 140 & 64.9 & - \\
RexNeXt-15 \cite{xie2017aggregated} & 140 & 65.7 & - \\
352-MENet-12$\times$1 ($g=8$, \emph{ours}) & 144 & \textbf{66.7} & \textbf{86.9} \\ \hline
ResNet-15 \cite{he2016deep} & 38 & 51.1 & - \\
Xception-15 \cite{chollet2016xception} & 38 & 53.9 & - \\
RexNeXt-15 \cite{xie2017aggregated} & 38 & 53.7 & - \\
108-MENet-8$\times$1 ($g=3$, \emph{ours}) & 38 & \textbf{56.1} & \textbf{79.2} \\ \hline
\end{tabular}
The results that surpasses all competing networks are \textbf{bold}.
Larger number in top-1 and top-5 accuracy represents better performance.
\end{threeparttable}
\end{table}

\begin{table}[!t]
\centering
\setlength{\tabcolsep}{2.5pt}
\begin{threeparttable}[!t]
\caption{ILSVRC 2012 Accuracy (\%) Comparison with ShuffleNet}
\label{table:result-shufflenet}
\begin{tabular}{|l|c|c|c|}
\hline
\textbf{Models} & \textbf{MFLOPs} & \textbf{Top-1 Acc.} & \textbf{Top-5 Acc.} \\ \hline
ShuffleNet 1$\times$ ($g=3$) \cite{zhang2017shufflenet} & 137 & 65.9 & - \\
ShuffleNet 1$\times$ ($g=4$) \cite{zhang2017shufflenet} & 134 & 65.7 & - \\
ShuffleNet 1$\times$ ($g=8$) \cite{zhang2017shufflenet} & 138 & 65.3 & - \\ \hline
ShuffleNet 1$\times$ ($g=3$, \emph{re-impl.}) \cite{zhang2017shufflenet} & 137 & 65.7 & 86.3 \\
ShuffleNet 1$\times$ ($g=4$, \emph{re-impl.}) \cite{zhang2017shufflenet} & 134 & 65.5 & 86.2 \\
ShuffleNet 1$\times$ ($g=8$, \emph{re-impl.}) \cite{zhang2017shufflenet} & 138 & 65.4 & 86.3 \\ \hline
228-MENet-12$\times$1 ($g=3$, \emph{ours}) & 144 & \textbf{66.4} & \textbf{86.7} \\
256-MENet-12$\times$1 ($g=4$, \emph{ours}) & 140 & \textbf{66.6} & \textbf{86.7} \\
352-MENet-12$\times$1 ($g=8$, \emph{ours}) & 144 & \textbf{66.7} & \textbf{86.9} \\ \hline
\end{tabular}
\end{threeparttable}
\end{table}

\begin{table}[!t]
\centering
\begin{threeparttable}[!t]
\setlength{\tabcolsep}{4.5pt}
\caption{ILSVRC 2012 Accuracy (\%) Comparison with ShuffleNet and MobileNet}
\label{table:results-compact-networks}
\begin{tabular}{|l|c|c|c|}
\hline
\textbf{Models} & \textbf{MFLOPs} & \textbf{Top-1 Acc.} & \textbf{Top-5 Acc.} \\ \hline
1$\times$ MobileNet-224 \cite{howard2017mobilenets} & 569 & 70.73 & 89.72 \\
ShuffleNet 2$\times$ ($g = 3$) \cite{zhang2017shufflenet} & 524 & 70.79 & 89.80 \\
456-MENet-24$\times$1 ($g = 3$, \emph{ours}) & 551 & \textbf{71.60} & \textbf{90.07} \\ \hline
0.75$\times$ MobileNet-224 \cite{howard2017mobilenets} & 325 & 68.60 & 88.33 \\
ShuffleNet 1.5$\times$ ($g = 3$) \cite{zhang2017shufflenet} & 292 & 68.85 & 88.41 \\
348-MENet-12$\times$1 ($g=3$, \emph{ours}) & 299 & \textbf{69.91} & \textbf{89.08} \\ \hline
0.5$\times$ MobileNet-224 \cite{howard2017mobilenets} & 149 & 64.74 & 85.63 \\
ShuffleNet 1$\times$ ($g = 3$) \cite{zhang2017shufflenet} & 137 & 65.69 & 86.29 \\
228-MENet-12$\times$1 ($g = 3$, \emph{ours}) & 144 & 66.43 & 86.72 \\
352-MENet-12$\times$1 ($g = 8$, \emph{ours}) & 144 & \textbf{66.69} & \textbf{86.92} \\ \hline
0.25$\times$ MobileNet-224 \cite{howard2017mobilenets} & 41 & 53.71 & 77.04 \\
ShuffleNet 0.5$\times$ ($g = 3$) \cite{zhang2017shufflenet} & 38 & 55.40 & 78.70 \\
108-MENet-8$\times$1 ($g = 3$, \emph{ours}) & 38 & \textbf{56.08} & \textbf{79.24} \\ \hline
\end{tabular}
\end{threeparttable}
\end{table}

\subsection{ImageNet Classification}

The ILSVRC 2012 dataset is composed of a training set of 1.28 million images and a validation set of 50,000 images, which are categorized into 1,000 classes.
We train the networks on the training set and report the top-1 and the top-5 accuracy rates on the validation set using center-crop evaluations.

\subsubsection{Implementation Details}
\label{section:imagenet-details}

All our experiments are conducted using PyTorch \cite{collobert2011torch7} with four GPUs.
We utilize synchronous stochastic gradient descent to train the models for 120 epochs with a batch size of 256 and a momentum of 0.9.
Following \cite{zhang2017shufflenet}, a relatively small weight decay of 4e-5 is used to avoid underfitting.
The learning rate starts from 0.1, and is divided by 10 every 30 epochs.
Because our models are relatively small, we use less aggressive multi-scale data augmentation.
Color jittering is not adopted because we find it can lead to underfitting.
On evaluation, each validation image is first resized with its shorter edge to 256 pixels, and then evaluated using the center $224 \times 224$ pixels crop.

\subsubsection{Comparison with Other State-of-the-art Networks}

Table~\ref{table:result-state-of-the-art} demonstrates the comparison of MENet and some state-of-the-art network structures on ILSVRC 2012 dataset.

We first compare MENet with two popular networks, GoogLeNet \cite{szegedy2015going} and VGG-16 \cite{simonyan2014very}.
GoogLeNet provides 69.8\% top-1 accuracy and 89.6\% top-5 accuracy, while VGG-16 produces remarkably better top-1 accuracy of 71.5\%.
However, they are all computationally intensive.
In comparison, 456-MENet-24$\times$1 achieves 71.6\% top-1 accuracy and 90.2\% top-5 accuracy under a complexity of about 550 MFLOPs.
MENet significantly surpasses GoogLeNet by 1.8\% on top-1 accuracy with 2.8$\times$ fewer FLOPs, and slightly outperforms VGG-16 ($\sim$0.1\%) with 27$\times$ fewer FLOPs.

We further compare the ME module with the building structures of three state-of-the-art networks, ResNet \cite{he2016deep}, Xception \cite{chollet2016xception} and ResNeXt \cite{xie2017aggregated}.
Following \cite{zhang2017shufflenet}, we replace the ME modules with other structures in the architecture shown in Table~\ref{table:menet-architecture} and adapting the number of channels to the computational budgets.
These networks are referred to as ResNet-15, Xception-15 and RexNeXt-15\footnote{The number 15 indicates the number of building blocks in the network.} and use the results reported in \cite{zhang2017shufflenet} for comparison.
As shown in Table~\ref{table:result-state-of-the-art}, 352-MENet-12$\times$1 achieves significant improvements of 5.4\% over ResNet-15, 1.8\% over Xception-15 and 1\% over ResNeXt-15 under a complexity of 140 MFLOPs, while 108-MENet-8$\times$1 achieves improvements of 5.0\%, 2.2\% and 2.3\% under 40 MFLOPs, respectively.
These improvements have proven the effectiveness of ME modules in building compact networks.

\subsubsection{Comparison with Other Compact Networks}

\begin{figure}[!t]
\centering
\includegraphics[width=0.45\textwidth]{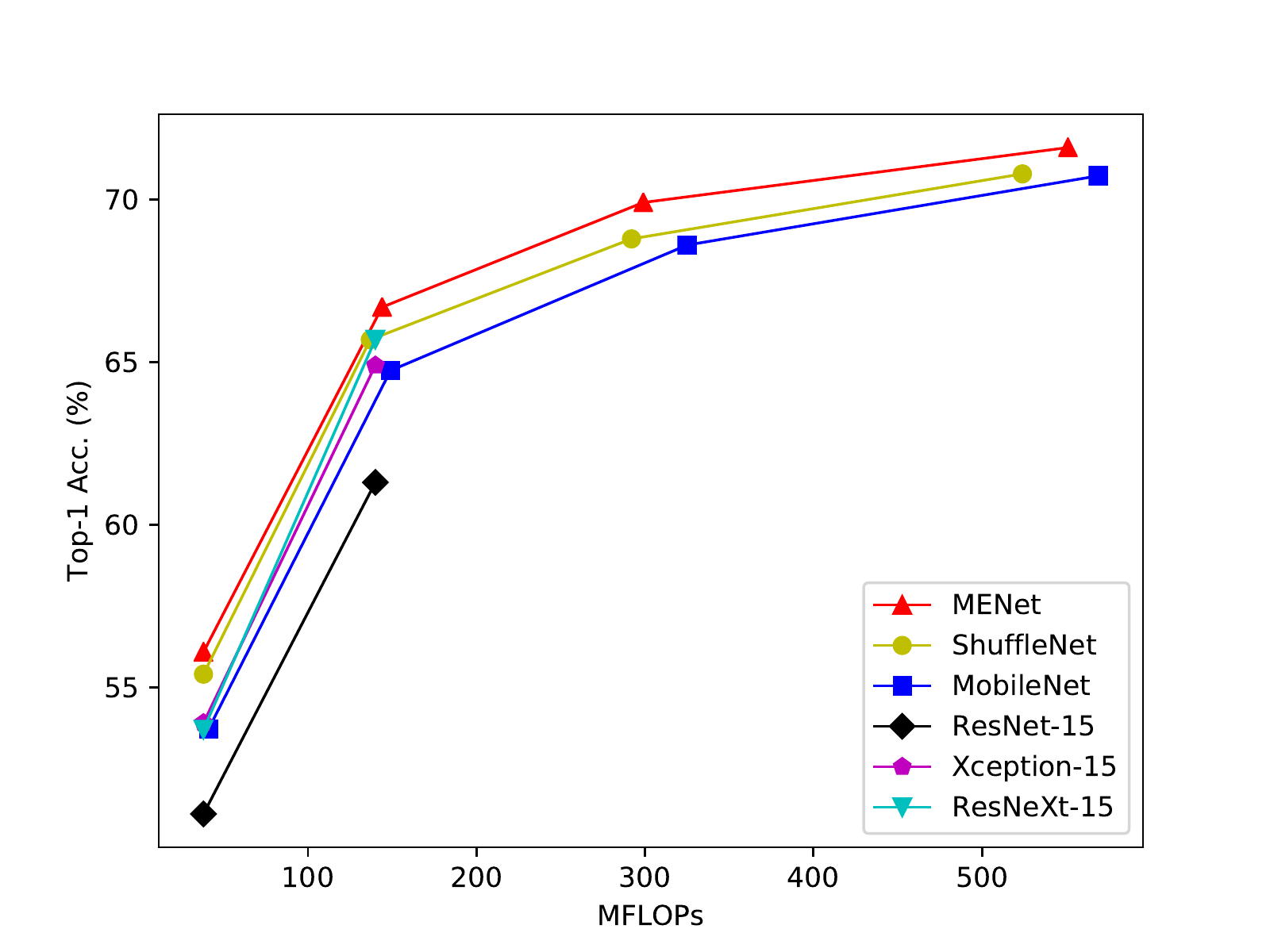}
\caption{Comparison of MENet with other network structures.
MENet surpasses all competing structures under all four computational budgets.}
\label{figure:results-all}
\end{figure}

We also compare the performance of MENet with two state-of-the-art compact networks: ShuffleNet \cite{zhang2017shufflenet} and MobileNet \cite{howard2017mobilenets}.
For fair comparison, we re-implement ShuffleNet and MobileNet with the same settings as described in Section~\ref{section:imagenet-details}.

Table~\ref{table:result-shufflenet} demonstrates the comparison of the results between MENet and ShuffleNet with different group numbers.
When the group number is kept the same, MENet surpasses ShuffleNet by a large margin.
Considering there are fewer channels in the residual branch in MENet than in ShuffleNet, we attribute this improvement to the effectiveness of the proposed merging and evolution operations.
Although ShuffleNet has more channels, it suffers from the loss of inter-group information.
On the other side, MENet leverages the inter-group information through the merging and evolution operations.
Consequently, MENet generates more discriminative features than ShuffleNet and overcomes the performance degradation.
This is the first advantage of MENet: \emph{it can achieve better performance with fewer channels}.

\begin{table}[!t]
\centering
\caption{ILSVRC 2012 Accuracy (\%) of Different Fusion Widths}
\label{table:result-fusion-width}
\begin{tabular}{|l|c|c|c|}
\hline
\textbf{Models} & \textbf{MFLOPs} & \textbf{Top-1 Acc.} & \textbf{Top-5 Acc.} \\ \hline
228-MENet-10$\times$1 ($g=3$) & 140 & 66.37 & 86.70 \\
228-MENet-12$\times$1 ($g=3$) & 144 & 66.43 & 86.72 \\
228-MENet-14$\times$1 ($g=3$) & 148 & 66.56 & 86.95 \\
228-MENet-16$\times$1 ($g=3$) & 153 & 66.86 & 87.19 \\ \hline
\end{tabular}
\end{table}

\begin{table}[!t]
\centering
\caption{ILSVRC 2012 Accuracy (\%) of Different Expansion Rates}
\label{table:result-expansion-rate}
\begin{tabular}{|l|c|c|c|c|}
\hline
\textbf{Models} & \textbf{MFLOPs} & \textbf{Top-1 Acc.} & \textbf{Top-5 Acc.} \\ \hline
228-MENet-12$\times$1 ($g=3$) & 144 & 66.43 & 86.72 \\
228-MENet-12$\times$1.5 ($g=3$) & 152 & 66.71 & 87.14 \\
228-MENet-12$\times$2 ($g=3$) & 163 & 67.25 & 87.30 \\
228-MENet-12$\times$2.5 ($g=3$) & 179 & 67.51 & 87.66 \\ \hline
\end{tabular}
\end{table}

\begin{table}[!t]
\centering
\setlength{\tabcolsep}{5pt}
\caption{ILSVRC 2012 Accuracy (\%) of Element-wise Product and Element-wise Addition}
\label{table:result-product-addition}
\begin{tabular}{|l|c|c|}
\hline
\textbf{Models} & \textbf{Top-1 Acc. (\%)} & \textbf{Top-5 Acc. (\%)} \\ \hline
228-MENet-12$\times$1 ($g=3$) & \textbf{66.43} & \textbf{86.77} \\
228-MENet-12$\times$1 (add, $g=3$) & 66.27 & 86.25 \\ \hline
256-MENet-12$\times$1 ($g=4$) & \textbf{66.53} & \textbf{86.82} \\
256-MENet-12$\times$1 (add, $g=4$) & 65.48 & 86.30 \\ \hline
\end{tabular}
\end{table}

In ShuffleNet, the top-1 accuracy decreases as the number of groups increases.
As figured out in Section~\ref{section:merging-evolution-operation}, the ratio of the inter-group connections lost is $\frac{G-1}{G}$ when there are $G$ channel groups.
Increasing $G$ makes more inter-group connections lost, which aggravates the loss of inter-group information.
More specifically, although there are more channels in total in the residual branch when the group number is larger, the number of channels within each channel group become smaller, which harms the representation capability.
However, the results are opposite for MENet: the classification accuracy rises given more channel groups.
This is another advantage that MENet brings: \emph{it can gain accuracy improvement by directly increasing the width of the network and the number of groups}.
The merging and evolution operations fuse the features from all channels simultaneously, thus alleviates the loss of inter-group information.
Consequently, MENet benefits from the wider feature maps and generates more discriminative features.
These improvements are consistent with our initial motivation to design ME modules.


\begin{table*}[!t]
\centering
\scriptsize
\setlength{\tabcolsep}{2.5pt}
\caption{Comparison of mAP (\%) and AP (\%) on PASCAL VOC 2007 Test Set}
\label{table:results-pascal-voc}
\begin{tabular}{|l|c|c|c|c|c|c|c|c|c|c|c|c|c|c|c|c|c|c|c|c|c|}
\hline
\textbf{Backbone} & \textbf{mAP} & \textbf{areo} & \textbf{bike} & \textbf{bird} & \textbf{boat} & \textbf{bottle} & \textbf{bus} & \textbf{car} & \textbf{cat} & \textbf{chair} & \textbf{cow} & \textbf{table} & \textbf{dog} & \textbf{horse} & \textbf{mbike} & \textbf{person} & \textbf{plant} & \textbf{sheep} & \textbf{sofa} & \textbf{train} & \textbf{tv} \\ \hline
0.5$\times$ MobileNet-224 & 54.8 & 59.4 & \textbf{65.4} & 52.7 & 36.3 & 30.1 & 57.2 & 71.3 & 64.4 & 31.1 & 64.4 & 47.5 & 62.5 & 70.8 & 68.8 & 64.2 & 27.6 & 53.2 & 51.8 & 61.8 & 54.9 \\
ShuffleNet 1$\times$ ($g=3$) & 56.6 & 57.4 & 64.6 & 55.4 & 37.7 & 30.4 & 60.1 & 72.1 & 66.8 & 29.9 & 64.0 & 50.3 & \textbf{65.7} & 73.5 & 68.6 & 66.9 & 34.2 & 58.5 & 50.0 & 65.7 & 60.0 \\
352-MENet-12$\times$1 ($g = 8$) & \textbf{58.9} & \textbf{64.2} & 64.1 & \textbf{59.0} & \textbf{44.6} & \textbf{34.6} & \textbf{64.1} & \textbf{73.8} & \textbf{70.0} & \textbf{32.8} & \textbf{69.1} & \textbf{52.5} & 64.3 & \textbf{75.3} & \textbf{69.4} & \textbf{67.7} & \textbf{35.3} & \textbf{59.5} & \textbf{50.6} & \textbf{66.9} & \textbf{60.5} \\ \hline
1$\times$ MobileNet-224 & 62.4 & 65.1 & 69.0 & 59.9 & 51.4 & 40.1 & 66.2 & 75.7 & 76.7 & 40.0 & 69.3 & 52.9 & 72.9 & 75.2 & 68.3 & 71.2 & 35.3 & 64.8 & 58.1 & 70.0 & 66.8 \\
ShuffleNet 2$\times$ ($g=3$) & 63.5 & 68.4 & 69.7 & 58.8 & 49.7 & 38.2 & 70.5 & 76.4 & 77.0 & 38.7 & 73.0 & 54.7 & 73.8 & \textbf{79.0} & 71.1 & 72.0 & 36.1 & 65.8 & \textbf{60.2} & 71.4 & 64.7 \\
456-MENet-24$\times$1 ($g=3$) & \textbf{65.5} & \textbf{69.2} & \textbf{72.6} & \textbf{66.5} & \textbf{52.1} & \textbf{42.3} & \textbf{70.8} & \textbf{79.4} & \textbf{78.9} & \textbf{41.3} & \textbf{75.7} & \textbf{56.3} & \textbf{78.0} & 77.4 & \textbf{71.9} & \textbf{74.4} & \textbf{38.0} & \textbf{66.8} & 58.2 & \textbf{72.2} & \textbf{67.4} \\ \hline
\end{tabular}
\end{table*}


We further compare the three compact networks under four computational budgets.
The results are demonstrated in Table~\ref{table:results-compact-networks}.
The number of output channels in the first convolution in MENet is adjusted to fit the computational budget.
According to the table, MENet significantly outperforms ShuffleNet and MobileNet under all the computational budgets.
Under a budget of 140 MFLOPs, MENet surpasses ShuffleNet by 0.74\% on top-1 accuracy with the same group number ($g=3$), and by 1\% with more groups ($g=8$).
Meanwhile, MENet surpasses MobileNet by 1.95\%.
We do not tune too much on group numbers for MENet but simply set $g=3$ under other computational budgets.
For smaller networks with only 40 MFLOPs, MENet provides improvements of 0.68\% over ShuffleNet and 2.37\% over MobileNet.
For larger networks under a complexity of 300 MFLOPs, MENet performs 1.06\% better than ShuffleNet and 1.31\% better than MobileNet.
When the complexity is 550 MFLOPs, MENet surpasses ShuffleNet and MobileNet by 0.81\% and 0.87\% respectively.
Similar results are observed on the top-5 accuracy.
More detailed comparison results are illustrated in Fig.~\ref{figure:results-all}.
These results have proven that MENet has stronger representation capability and is both efficient and accurate for various scenarios.

\subsubsection{Model Choices}

We furthermore conduct experiments to examine the influences of several design choices on the performance of MENet, including the fusion width, the expansion rate, and the function which combines the fusion branch and the residual branch.

\textbf{Fusion Width}.
The fusion width is the hyper-parameter which controls the initial number of channels in the fusion branch.
We evaluate the effects of the fusion width using four models: 228-MENet-10$\times$1, 228-MENet-12$\times$1, 228-MENet-14$\times$1 and 228-MENet-16$\times$1, all with $g=3$.
Table~\ref{table:result-fusion-width} shows the comparison of these networks.
Substantial improvements in accuracy are observed as the fusion width increases.
In ME modules, we set the fusion branch to be relatively narrow for computational efficiency.
This limits the representation capability of the features generated from the fusion branch.
Increasing the fusion width improves the information capacity of the fusion branch, which allows more inter-group information to be encoded and improves the representation capability.

\textbf{Expansion Rate}.
The expansion rate controls the ``growth'' of the channels in the fusion branch between stages.
We also select four MENet models to examine the effect of the expansion rate: 228-MENet-12$\times$1, 228-MENet-12$\times$1.5, 228-MENet-12$\times$2, 228-MENet-12$\times$2.5, all with $g=3$.
The results are shown in Table~\ref{table:result-expansion-rate}.
It is observed that the networks with larger expansion rates are inclined to have higher accuracy.
The model with an expansion rate of 2.5 achieves an improvement above 1\% on top-1 accuracy over the model whose expansion rate is 1.
It is conjectured that as the width of the residual branch increases from stage to stage, the inter-group information becomes increasingly complicated.
This makes it difficult to encode all the information within a fixed number of channels in the fusion branch for all stages.
By applying a large expansion rate, different number of channels are used to fuse the features in each stage, which helps improve the representation capability in the later stages.

\textbf{Element-wise Product \textit{vs.} Element-wise Addition}.
It is a conventional practice to learn residual information (element-wise addition) in state-of-the-art deep networks \cite{he2016deep, xie2017aggregated, szegedy2017inception}.
However, we choose to learn neuron-wise scaling information (element-wise product) instead in MENet.
We evaluate the effects of these two choices using two MENet models with different group numbers ($g=3$ and $g=4$).
For the networks using element-wise addition, we simply make two modifications:
(i) The element-wise product is replaced by an element-wise addition.
(ii) The sigmoid activation after the second pointwise convolution in the fusion branch is removed.
The results are demonstrated in Table~\ref{table:result-product-addition}.
It is clear that learning scaling information significantly outperforms residual information.
The model with element-wise product is 0.16\% better when $g=3$, and 1.05\% better when $g=4$.
Notice that the model learning residual information provides a worse result than its ShuffleNet counterpart when $g=4$.
These results indicate that residual information is not effective for inter-group feature fusion.
This difference may be potentially induced by the narrow feature maps in the fusion branch, which cannot encode adequate residual information.
We are planning to further examine this in our future work.

\subsection{Object Detection on PASCAL VOC}

To investigate the generalization ability of MENet, we conduct comparative experiments on PASCAL VOC 2007 detection dataset \cite{everingham2010pascal}.
PASCAL VOC 2007 dataset consists of about 10,000 images split into three (train/val/test) sets.
We train the models on VOC 2007 trainval set and report the single-model results on VOC 2007 test set.


We adopt Faster R-CNN \cite{ren2015faster} detection pipeline and compare the performance of MENet, ShuffleNet and MobileNet on 600$\times$ resolution under two computational budgets (140 MFLOPs and 550 MFLOPs).
The pre-trained models on ILSVRC 2012 dataset are used for transfer learning.
For the MobileNet-based detectors, we use the first 28 layers as the R-CNN base network and the remaining 4 layers as the R-CNN subnet.
For the ShuffleNet-based and the MENet-based detectors, the first three stages are used as the base network, and the last stage is used as the R-CNN subnet.
All strides in R-CNN subnets are set 1 to obtain larger feature maps.
RoI align \cite{he2017mask} is used to encode RoIs instead of RoI pooling \cite{girshick2015fast}.
During testing, 300 region proposals are sent to the R-CNN subnet to generate the final predictions.



Table~\ref{table:results-pascal-voc} demonstrates the comparison of the three compact networks on VOC 2007 test set.
According to the results, MENet significantly outperforms MobileNet and ShuffleNet under both computational budgets.
Under the computational budget of 140 MFLOPs, the MENet-based detector achieves the mAP of 58.9\%, while the mAP of the ShuffleNet-based and the MobileNet-based detectors is 56.6\% and 54.8\%, respectively.
MENet achieves improvements of 2.3\% mAP over ShuffleNet and 4.1\% over MobileNet.
More specifically, MENet provides better results on most classes, with the improvements from 0.5\% (tv) to 6.9\% (boat).
On the classes which are difficult for ShuffleNet and MobileNet, such as boat, bottle, table and plant, the MENet-based detector increases the AP by 6.9\%, 4.2\%, 2.2\% and 1.1\%, respectively.
Under a complexity of 550 MFLOPs, the MENet-based detector surpasses the ShuffleNet-based one by 2\% and the MobileNet-based one by 3.1\% on mAP.
Additionally, MENet also outperforms ShuffleNet and MobileNet on single-class results.
These results have proven that the proposed MENet has strong generalization ability and can benefit various tasks.

\section{Conclusion}

In this paper, we propose two novel operations, merging and evolution, to perform feature fusion across all channels in a group convolution and alleviate the performance degradation induced by the loss of inter-group information.
Based on the proposed operations, we introduce an architectural unit named ME module specially designed for compact networks.
Finally, we propose MENet, a family of compact neural networks.
Compared with ShuffleNet, the proposed MENet leverages inter-group information and generates more discriminative features.
Extensive experiments show that MENet consistently outperforms other state-of-the-art compact neural networks under different computational budgets.
Experiments on object detection show that MENet has strong generalization ability for transfer learning.
For future work, we consider to further evaluate MENet on other tasks such as semantic segmentation.

\section*{Acknowledgment}

This work is supported by the National Key Research and Development Program of China (2016YFB1000100).

\bibliographystyle{IEEEtran}
\bibliography{main}

\end{document}